\pdfoutput=1

\documentclass[11pt]{article}
\usepackage{booktabs}
\usepackage{multirow}
\usepackage{enumitem}
\usepackage[table]{xcolor}

\usepackage[final]{acl}
\usepackage{tabularx}
\usepackage[table]{xcolor}
\usepackage{subcaption}

\usepackage{times}
\usepackage{latexsym}
\usepackage{placeins}
\usepackage[T1]{fontenc}

\usepackage[utf8]{inputenc}

\usepackage{microtype}

\usepackage{inconsolata}
\usepackage{makecell}

\usepackage{graphicx}

\usepackage[final]{changes} 
\definechangesauthor[name={Joyeeta}, color=blue]{JD}

\title{Exploring the Limits of Model Compression in LLMs:\\ A Knowledge Distillation Study on QA Tasks}

\author{
 \textbf{Joyeeta Datta\textsuperscript{1}},
 \textbf{Niclas Doll\textsuperscript{1}},
 \textbf{Qusai Ramadan\textsuperscript{2}},
 \textbf{Zeyd Boukhers\textsuperscript{3,4}}
\\
 \textsuperscript{1}Fraunhofer Institute for Intelligent Analysis and Information Systems IAIS, Germany \\
 \textsuperscript{2}University of Southern Denmark, Denmark \\
 \textsuperscript{3}Fraunhofer Institute for Applied Information Technology FIT, Germany \\
 \textsuperscript{4}University Hospital of Cologne, Germany
\\
\texttt{\{\href{mailto:joyeeta.datta@iais.fraunhofer.de}{joyeeta.datta}, \href{mailto:niclas.doll@iais.fraunhofer.de}{niclas.doll}\}@iais.fraunhofer.de}\\
\texttt{\href{mailto:qura@mmmi.sdu.dk}{qura@mmmi.sdu.dk}}\\
\texttt{\href{mailto:zeyd.boukhers@fit.fraunhofer.de}{zeyd.boukhers@fit.fraunhofer.de}}
\\
\\
}

\begin{document}
\maketitle
\begin{abstract}
Large Language Models (LLMs) have demonstrated outstanding performance across a range of NLP tasks; however, their computational demands hinder their deployment in real-world, resource-constrained environments. This work investigates the extent to which LLMs can be compressed using Knowledge Distillation (KD) while maintaining strong performance on Question Answering (QA) tasks. We evaluate student models distilled from the Pythia and Qwen2.5 families on two QA benchmarks, SQuAD and MLQA, under zero-shot and one-shot prompting conditions. Results show that student models retain over 90\% of their teacher models’ performance while reducing parameter counts by up to 57.1\%. Furthermore, one-shot prompting yields additional performance gains over zero-shot setups for both model families. These findings underscore the trade-off between model efficiency and task performance, demonstrating that KD, combined with minimal prompting, can yield compact yet capable QA systems suitable for resource-constrained applications.
\end{abstract}

\section{Introduction}

In recent years, the increasing capabilities of Large Language Models (LLMs) have significantly advanced the field of NLP, enabling state-of-the-art results in tasks ranging from Question Answering (QA) \citep{brown2020language} to summarization \citep{raffel2023exploringlimitstransferlearning} and translation \citep{lewis2019bartdenoisingsequencetosequencepretraining}.
However, their considerable memory and compute demands hinder practical deployment, particularly in resource-constrained environments \citep{howell2023economictradeoffslargelanguag} \citep{kaddour2023challengesapplicationslargelanguage}. These challenges have fueled interest in model compression techniques, which aim to reduce model size while preserving the performance of large models.

Among these methods, Knowledge Distillation (KD) \citep{hinton2015distilling} has emerged as a widely adopted strategy for transferring the behavior of a larger teacher model to a smaller student model. While prior studies have demonstrated its effectiveness across various NLP tasks, it remains unclear to what extent the compression is possible without degrading task-specific performance.

This work studies the trade-offs between model size and task performance by distilling models of varying sizes from the \emph{Pythia}~\citep{pythia} and \emph{Qwen2.5}~\citep{qwen2025qwen25technicalreport} families. To complement this scaling analysis, we evaluate how lightweight prompting strategies, i.e., zero-shot and one-shot prompting, affect model behavior. Our contributions are as follows: (1) We perform a scaling analysis by distilling multiple student models from larger teacher models and evaluating their performance on QA tasks, (2) We assess these models under zero-shot and one-shot prompting conditions, offering insights into the role of lightweight prompting in compressed models.

\section{Related Work}
KD has proven effective across a variety of NLP tasks, including classification and sequence generation, with notable examples such as DistilBERT~\citep{sanh2020distilbert} and TinyBERT \citep{jiao2020tinybert}. For autoregressive models, the DISTILLM framework \citep{ko2024distillm} proposed Skew Kullback-Leibler Divergence and off-policy sampling strategies to improve stability and distillation quality.

Our work builds directly on DISTILLM by applying its distillation methodology to the Pythia \citep{pythia} and Qwen2.5 \citep{qwen2025qwen25technicalreport} model families.

In parallel, few-shot prompting, a form of In-Context Learning (ICL), allows models to generalize to new tasks by conditioning on a small number of examples in the input prompt \citep{brown2020language}. Recent work has begun to explore the intersection of KD and ICL. Notably, \citet{huang2022incontextlearningdistillationtransferring} introduced in-context learning distillation to transfer few-shot capabilities from larger to smaller models, focusing on meta-trained models evaluated on multitask and instruction-tuned benchmarks, CrossFit \citep{ye2021crossfitfewshotlearningchallenge} and LAMA \citep{petroni-etal-2019-language}. Similarly, \citet{sauer-etal-2022-knowledge} examined KD for few-shot intent classification.

Our study differs in two key ways: (1) we focus on extractive QA tasks, rather than multitask or classification settings, and (2) we explicitly evaluate performance under zero-shot and one-shot prompting, without relying on instruction tuning or prompt mixture training. This setup allows us to isolate how prompt structure and model size interact, offering new insights into the compression-performance trade-off in QA settings.

\section{Methodology}

We conduct a scaling analysis using distilled models of varying sizes and assess their performance under different prompting strategies. 

\subsection{Data Preprocessing}
We conduct our experiments on two widely used QA benchmarks: SQuAD v2.0  \citep{rajpurkar2018squad2} and MLQA \citep{mlqa}. 

SQuAD v2.0 contains over 150K English QA pairs, including a subset of unanswerable questions, which allows us to assess model performance in ambiguous scenarios. MLQA is a multilingual benchmark covering seven languages. In this work, we focus on the English and German subsets to evaluate cross-lingual generalization capabilities.

To study ICL capabilities of the models, we create zero-shot and one-shot variants of each dataset. In the zero-shot setting, each model sees only the input context and question. In the one-shot setting, we prepend a single in-context example to each instance. To avoid overlap, the demonstration set used for constructing one-shot prompts is excluded from both the training and test sets.

Table~\ref{tab:dataset-format} in the appendix illustrates the dataset formats for both prompting setups.

\subsection{Prompting Strategy}
We used simple few-shot prompting by prepending a single representative example to every instance in the dataset. The prompt format follows a standard QA structure, comprising a context, a question, and an answer, and was kept consistent across all models and datasets to ensure comparability. While more advanced prompting strategies (e.g., chain-of-thought) could yield better performance, we focused on isolating the effect of model compression by applying minimal and uniform prompting across all settings, ensuring a controlled experimental setup.

\subsection{Model and Dataset Selection}
We selected the Pythia \citep{pythia} and Qwen2.5 \citep{qwen2025qwen25technicalreport} model families due to their open-source availability, diverse range of model sizes, and strong baseline performance on QA tasks. During this study, these were among the few publicly available model families with multiple size variants under a unified architecture, making them well-suited for studying model compression trade-offs. Also, both are compatible with the DISTILLM framework, facilitating efficient knowledge distillation and evaluation.

We chose SQuAD v2.0 and MLQA datsets as they represent two widely adopted QA benchmarks with complementary characteristics. SQuAD provides high-quality span-based annotations in English, including a substantial portion of unanswerable questions, which allows for evaluating model robustness under ambiguity. MLQA extends the evaluation to multilingual settings, offering parallel QA examples across multiple languages. We focus on the English and German subsets to assess cross-lingual generalization. Together, these models and datasets form a comprehensive experimental setup for analyzing the effectiveness of knowledge distillation in diverse QA scenarios.

\subsection{Model Training and Distillation}

We distill student models from two teacher models: Qwen2.5-7B and Pythia-2.8B. Each teacher is fine-tuned on the corresponding dataset and prompting configuration (zero-shot and one-shot). KD is then performed using the DISTILLM framework \citep{ko2024distillm}, transferring the teacher's behavior to smaller student models. For Qwen2.5, student models range from 3B to 0.5B parameters, while for Pythia, they range from 1.4B down to 70M parameters. Full model configurations are provided in the Appendix section (Table~\ref{table:teacher_student_models}).

\subsection{Evaluation Setup}
We evaluate all models using Exact Match (EM) \citep{rajpurkar2016squad} and ROUGE-L \citep{ganesan2018rouge}, capturing both answer precision and semantic overlap. 
To ensure robustness, all experiments are run across five different random seeds, and results are reported as the average across these runs.

We compare:
(a) fine-tuned teacher models,
(b) distilled student models, and
(c) fine-tuned student models.

Specifically, we analyze:
(1) how model size impacts performance retention,
(2) the effect of prompting strategies on compressed models,
(3) the comparative strengths of distillation vs. fine-tuning.


\FloatBarrier

\section{Results and Discussion}
In this section, we present the empirical results from our experiments, analyzing the effects of model compression and prompting strategies on QA performance. 

\paragraph{Model Size vs. Performance:}

\begin{figure}[t]
  \includegraphics[width=\columnwidth]{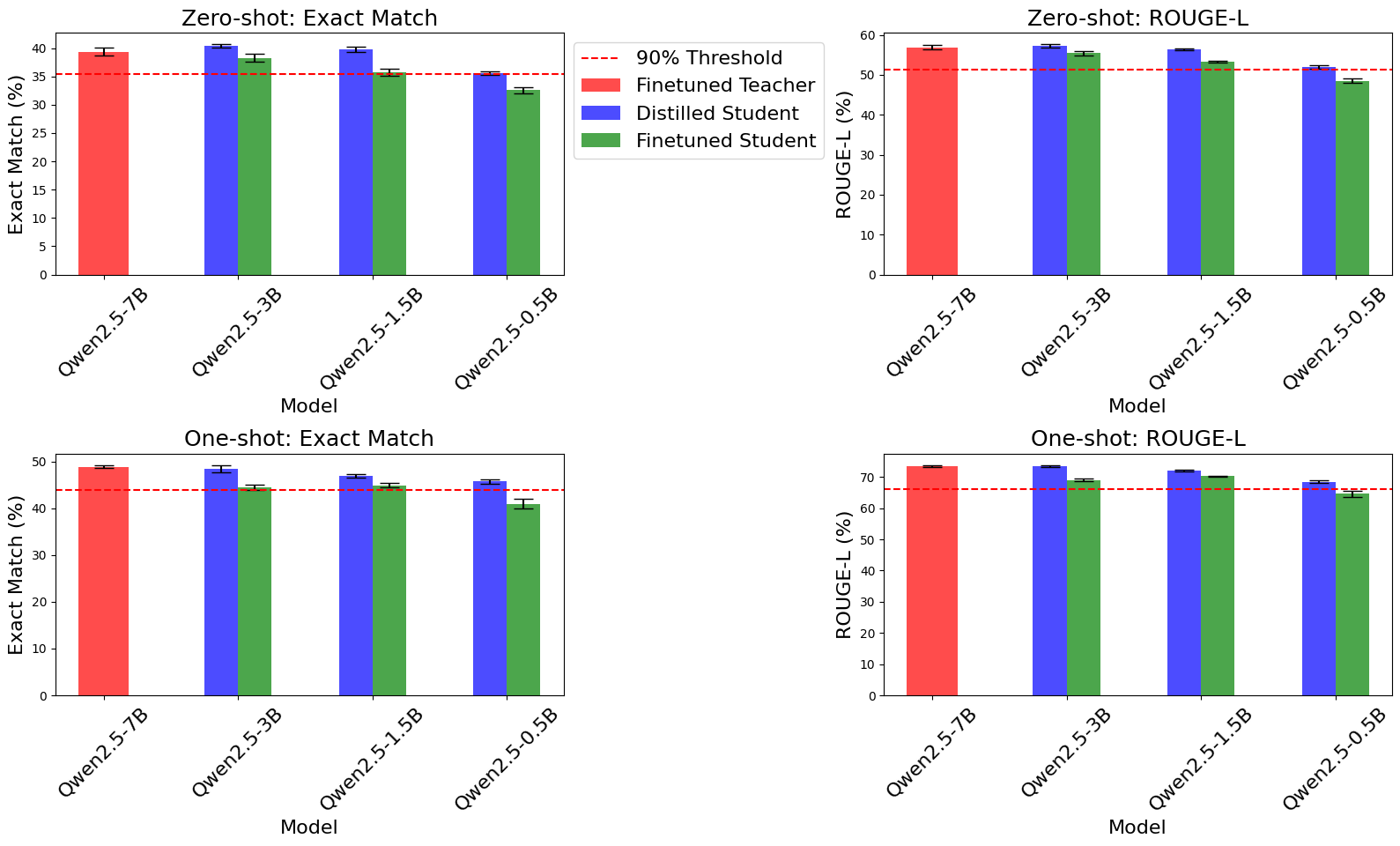}
  \caption{Exact Match and ROUGE-L scores for Qwen2.5 models on the MLQA English split under zero-shot and one-shot settings. Error bars denote standard deviation across five random seeds.}
  \label{fig:qwen_mlqa}
\end{figure}

\begin{figure}[t]
    \centering
    \includegraphics[width=\columnwidth]{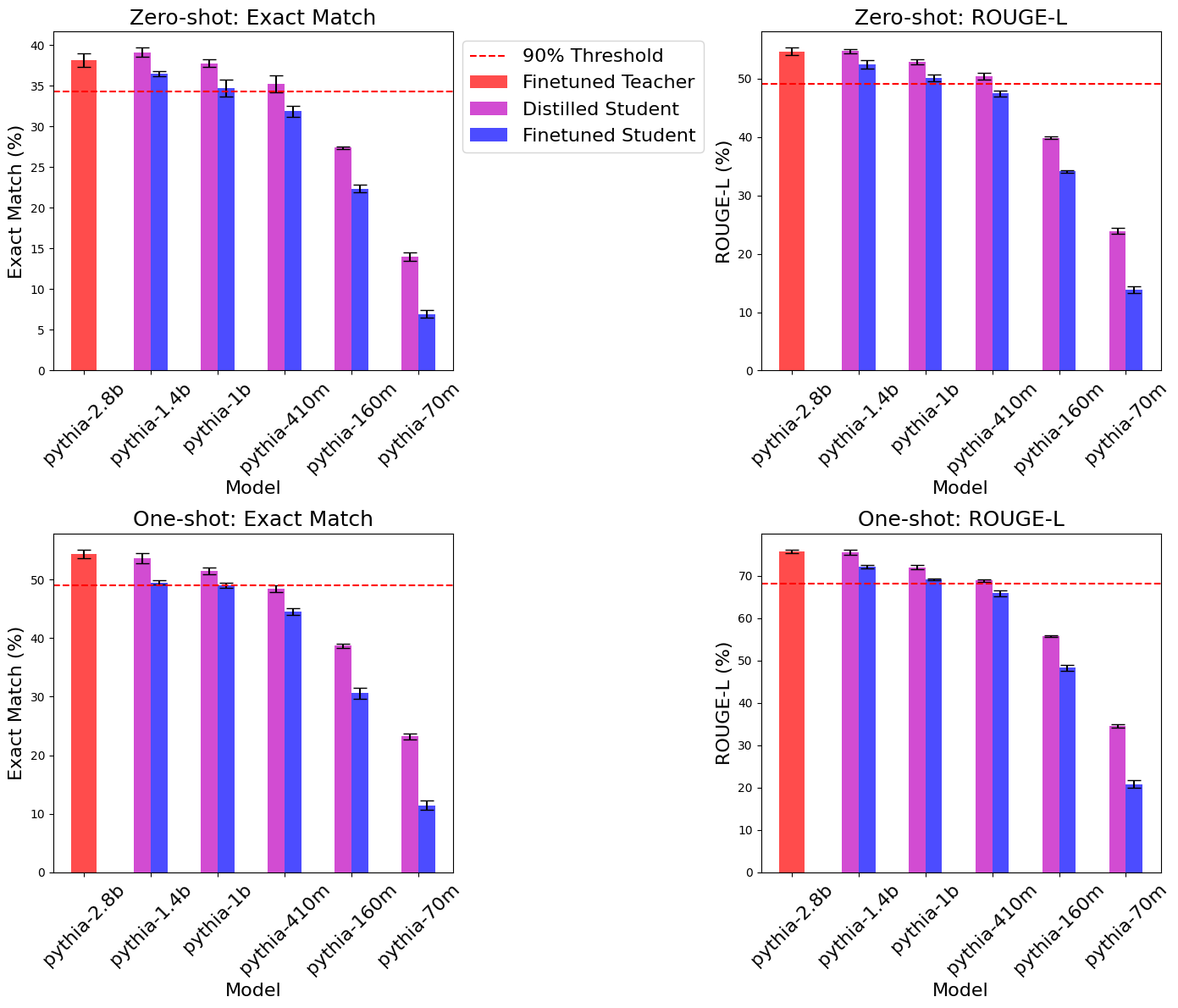}
     \caption{Exact Match and ROUGE-L scores for Pythia models on the MLQA English split under zero-shot and one-shot settings. Error bars denote standard deviation across five random seeds.}
    \label{fig:pythia_mlqa}
\end{figure}

Figures~\ref{fig:qwen_mlqa} and~\ref{fig:pythia_mlqa} illustrate how model performance scales with size across the Qwen2.5 and Pythia model families under both zero-shot and one-shot prompting conditions. We observe that larger student models, such as Qwen2.5-3B and Pythia-1.4B, consistently retain over 90\% of their teacher models' performance in both EM and ROUGE-L metrics. In particular, the distilled Qwen2.5-3B closely matches or even slightly exceeds the performance of the Qwen2.5-7B teacher in zero-shot settings, suggesting that moderate compression can enhance generalization.

As the model size decreases, performance degrades more noticeably, especially for smaller models like Qwen2.5-0.5B and Pythia-70M. This trend is consistent across both zero-shot and one-shot settings, suggesting that maintaining strong task-specific capabilities under compression becomes increasingly challenging at lower parameter scales. Nevertheless, moderately sized student models keep a good balance between efficiency and performance, making them good candidates for resource-constrained applications.

All experiments were run with five different random seeds, with results showing low variance and high reproducibility across runs (see Appendix~\ref{sec:var_anal} for detailed variance analysis).

\paragraph{Distillation vs. Fine-Tuning:}
\emph{Distilled models generally outperform their fine-tuned counterparts.} For instance, the distilled Qwen2.5-3B and Pythia-1.4B models achieve substantially higher scores compared to their fine-tuned counterparts, narrowing the gap to their teacher models. In contrast, fine-tuned students show larger performance degradation, particularly at smaller scales (e.g., Qwen2.5-0.5B and Pythia-70M). These results demonstrate that KD not only compresses models effectively but also enhances their ability to generalize from prompts, making it a more reliable strategy for scaling down LLMs without significant task-specific performance loss.

\paragraph{Effect of Prompting (Zero-shot vs. One-shot):}

We observe that one-shot prompting generally improves performance across most datasets and models. As shown in Table~\ref{tab:compact-summary}, student models like Qwen2.5-3B and Pythia-1.4B gain substantial EM and ROUGE-L improvements under one-shot prompting compared to zero-shot settings.

For instance, Qwen2.5-3B achieves a +5.4 EM point improvement on the MLQA German split, and Pythia-1.4B gains +14.5 EM points on the MLQA English split. These findings suggest that providing an in-context example can significantly enhance the generalization ability of compressed models.

Another key observation is the variability in performance retention across languages: Qwen2.5-3B performs robustly even on the MLQA German subset, showing stronger multilingual generalization than the Pythia models, which were evaluated only on English. These results highlight not only the feasibility of compression but also the importance of model family and language in shaping student model effectiveness.

\begin{table*}[t]
\centering
\small
\setlength{\tabcolsep}{3.5pt}
\begin{tabular}{|l|l|cc|cc|cc|cc|cc|cc|}
\hline
\multirow{3}{*}{\textbf{Dataset}} & \multirow{3}{*}{\textbf{Setting}} & \multicolumn{6}{c|}{\textbf{Qwen2.5 Models}} & \multicolumn{6}{c|}{\textbf{Pythia Models}} \\
\cline{3-14}
& & \multicolumn{2}{c|}{Teacher (7B)} & \multicolumn{4}{c|}{Best Student} & \multicolumn{2}{c|}{Teacher (2.8B)} & \multicolumn{4}{c|}{Best Student} \\
\cline{3-14}
& & EM & R-L & EM & R-L & Model & Size & EM & R-L & EM & R-L & Model & Size \\
\hline
\multirow{2}{*}{MLQA (EN)} 
& Zero & 39.32 & 56.92 & \textbf{40.33} & \textbf{57.28} & 3B & 43\% & 38.14 & 54.66 & \textbf{39.15} & \textbf{54.76} & 1.4B & 50\% \\
& One & 48.87 & 73.34 & 48.45 & 73.47 & 3B & 43\% & 54.36 & 75.72 & 53.65 & 75.53 & 1.4B & 50\% \\
\hline
\multirow{2}{*}{MLQA (DE)} 
& Zero & 26.25 & 43.19 & \textbf{28.32} & \textbf{44.38} & 3B & 43\% & - & - & - & - & - & - \\
& One & 29.17 & 55.21 & \textbf{32.91} & \textbf{57.87} & 3B & 43\% & - & - & - & - & - & - \\
\hline
\multirow{2}{*}{SQuAD} 
& Zero & 62.51 & 73.29 & \textbf{64.09} & \textbf{74.50} & 3B & 43\% & 59.78 & 70.46 & \textbf{59.86} & 69.72 & 1.4B & 50\% \\
& One & 65.03 & 78.37 & \textbf{65.23} & 78.05 & 3B & 43\% & 43.23 & 66.36 & 42.86 & 66.10 & 410M & 15\% \\
\hline
\end{tabular}
\caption{Top-performing student models retain $\geq90\%$ of their teacher’s performance under Exact Match (EM) and ROUGE-L (R-L) metrics. Bold values indicate where students outperform teachers. The size column shows the student model size as a percentage of the teacher model. The dash (-) indicates no experiments were conducted.}
\label{tab:compact-summary}
\end{table*}

\paragraph{Prompt Sensitivity and Evaluation Discrepancies:}
The impact of prompting is not uniformly positive across all models. Notably, we observed an inconsistency with Pythia models on the SQuAD dataset, where one-shot prompting improved performance during validation but led to significant performance degradation during evaluation. As shown in Table~\ref{tab:val_test_discrepancy}, the Pythia-1.4B model achieved substantially better performance with one-shot prompting during validation; however, on the test set, the same model significantly underperformed in the one-shot setting compared to zero-shot.

\begin{table}[th!]
\centering
\begin{tabular}{|l|l|c|c|}
\hline
\textbf{Split} & \textbf{Setting} & \textbf{EM} & \textbf{ROUGE-L} \\
\hline
\multirow{2}{*}{Validation} & Zero-shot & 63.60 & 73.18 \\
                           & One-shot  & \textbf{68.00} & \textbf{78.74} \\
\hline
\multirow{2}{*}{Test} & Zero-shot & \textbf{59.86} & \textbf{69.72} \\
                     & One-shot  & 43.23 & 66.36 \\
\hline
\end{tabular}
\caption{Validation vs. evaluation scores of Pythia-1.4B on the SQuAD dataset. Despite a strong one-shot performance during validation, the evaluation score drops significantly in the one-shot setting.}
\label{tab:val_test_discrepancy}
\end{table}

This discrepancy suggests that while the model could effectively leverage in-context examples during training and validation, it failed to generalize that behavior to the test set. Several factors may contribute to this phenomenon:

\begin{itemize}[leftmargin=*]
    \item \textbf{Prompt length sensitivity}: The longer prompts in one-shot settings may affect model behavior differently depending on the specifics of the validation vs. test examples.
    \item \textbf{Overfitting to few-shot structure}: The model may have overfitted to the specific structure or patterns in the validation examples.
    \item \textbf{Distribution shifts}: Subtle differences in example distributions between validation and test splits may affect how well in-context learning generalizes.
\end{itemize}

This finding highlights the importance of comprehensive evaluation setups when assessing few-shot learning capabilities and suggests that prompt engineering and robust validation procedures are crucial when deploying knowledge-distilled models in real-world scenarios.

\paragraph{Implications and Observations:}
This study yields several important observations about the relationship between model size, prompting strategies, and the effectiveness of KD in QA tasks.

First, our results demonstrate that KD enables student models to retain strong ICL capabilities. For example, Qwen2.5-3B and Pythia-1.4B consistently achieved over 90\% of their teacher models’ performance with significantly fewer parameters. This suggests that distillation is a practical strategy for compressing LLMs without substantially sacrificing performance, making these models more suitable for deployment in resource-constrained environments.

Second, we observe that the benefit of prompting strategies is not uniform across model sizes. Larger models, such as Qwen2.5-3B, benefit consistently from one-shot prompting, particularly on cross-lingual QA tasks like MLQA (German). In contrast, smaller models, such as Pythia-70M, struggle to utilize in-context examples effectively, even when overall variance is low. This suggests a capacity threshold, below which few-shot learning becomes less effective, highlighting the importance of considering model size when applying ICL.

Finally, we observe inconsistencies between validation and test performance in some cases, most notably for Pythia models on the SQuAD dataset. While one-shot prompting improved validation results, the same setup led to underperformance during evaluation. This points to a potential sensitivity to prompt structure or dataset distribution differences across splits. It highlights the need for careful prompt design and evaluation setup when assessing few-shot learning behavior.

Taken together, these findings suggest that while distillation offers a promising path toward model efficiency, the interaction between compression and ICL behavior needs further investigation, particularly in settings involving diverse prompting strategies such as instruction-based or chain-of-thought formats.

\section*{Limitations}
While this study offers valuable insights into compressing LLMs through KD under zero-shot and one-shot prompting, it also shows limitations that suggest avenues for future research. We focus only on simple prompting strategies based on single example concatenation, without exploring more advanced formats such as chain-of-thought or instruction-based prompting, which may further enhance ICL capabilities.

Additionally, we observe inconsistencies between validation and test performance, particularly for Pythia models on SQuAD, suggesting sensitivity to prompt structure and dataset splits. This points to potential issues with generalization and prompt sensitivity, which need further investigation—especially with respect to prompt design, dataset splits, and overfitting to seen examples.

Finally, although this work centers on compression via KD, combining it with other techniques such as quantization or parameter-efficient fine-tuning (e.g., LoRA, adapters) could further improve model deployability and efficiency.

\section{Conclusion}
This study offers a detailed investigation into compressing LLMs through KD, with a focus on retaining their ICL abilities. Evaluating models from the Pythia and Qwen2.5 families on both SQuAD and MLQA datasets, we find that distilled student models can retain over 90\% of their teacher models’ performance while achieving substantial reductions in parameter size. One-shot prompting further increases performance, particularly in multilingual settings like MLQA (German), highlighting the usefulness of few-shot learning in compressed models. Overall, our findings suggest that KD, when combined with few-shot prompting, offers a promising direction for building compact, generalizable, and cost-efficient language models suitable for real-world deployment.

While our study focuses on extractive QA tasks, the observed trends, particularly the strong performance retention of distilled models under minimal few-shot prompting, may generalize to other NLP tasks that can be framed as QA problems. Recent work on instruction-tuned models such as UnifiedQA \citep{khashabi2020unifiedqacrossingformatboundaries} and FLAN \citep{wei2022finetunedlanguagemodelszeroshot} has shown that a wide range of tasks, including sentiment analysis, text classification, and natural language inference, can be effectively reformulated using a QA-style input-output format. 
Moreover, both the Pythia and Qwen2.5 model families have demonstrated strong baseline performance on non-QA benchmarks, such as LAMBADA, PIQA, and WinoGrande for Pythia \citep{pythia}, and MMLU, BBH, and ARC-C for Qwen2.5 \citep{qwen2025qwen25technicalreport}, highlighting their versatility beyond QA. Therefore, our findings may hold relevance for broader instruction-style applications, especially under few-shot or zero-shot settings. Future work could explore whether similar compression and prompting strategies yield consistent gains across diverse task families.

\bibliography{references}

\appendix

\section{Appendix}
\label{sec:appendix}
\renewcommand{\thetable}{A\arabic{table}}
\setcounter{table}{0}  

\subsection{Dataset Examples}

We provide additional details on the format of zero-shot and one-shot prompts used in our experiments.
\begin{table}[th!]
\centering
\scriptsize
\resizebox{0.9\linewidth}{!}{
\begin{tabular}{|p{0.9cm} | p{6.9cm}|}
\hline
\multicolumn{2}{|c|}{\textbf{Zero-shot Dataset}} \\
\hline
\textbf{id} & d8b0801e5f6428a965bec3977ee2f0d86e6146a0 \\
\hline
\textbf{prompt} &
\begin{minipage}[t]{\linewidth}\raggedright
\textbf{Context:} 2001: Classic Bruce Willis: The Universal Masters Collection (Polygram Int’l, OCLC 71124889)\\
\textbf{Question:} In what year was the Classic Bruce Willis collection released?
\end{minipage} \\
\hline
\textbf{output} & 2001 \\
\hline
\multicolumn{2}{|c|}{\textbf{One-shot Dataset}} \\
\hline
\textbf{id} & d8b0801e5f6428a965bec3977ee2f0d86e6146a0 \\
\hline
\textbf{prompt} &
\begin{minipage}[t]{\linewidth}\raggedright
\textbf{Example:} \\
\textbf{Context:} In March 1010 his successor, Basil Mesardonites, disembarked with reinforcements and besieged the rebels in the city...\\
\textbf{Question:} When did Mesardonites leave?\\
\textbf{Answer:} March 1010\\[0.5em]
\textbf{Current Task:}\\
\textbf{Context:} 2001: Classic Bruce Willis: The Universal Masters Collection (Polygram Int’l, OCLC 71124889)\\
\textbf{Question:} In what year was the Classic Bruce Willis collection released?
\end{minipage} \\
\hline
\textbf{output} & 2001 \\
\hline
\end{tabular}
}
\caption{Preprocessed zero-shot and one-shot examples.}
\label{tab:dataset-format}
\end{table}

\subsection{Model Configurations and Architectures}
In our scaling analysis, we experimented with various model sizes to investigate the trade-off between parameter count and performance. For each teacher model, we created several student models with progressively decreasing parameter counts. Table \ref{table:teacher_student_models} provides the complete details of all teacher and student models used in our experiments. The Qwen2.5 student models range from 3B parameters (43\% of teacher size) down to 0.5B parameters (7\% of teacher size), while the Pythia student models range from 1.4B parameters (50\% of teacher size) down to 70M parameters (2.5\% of teacher size).

\begin{table}[th!]
\centering
\small
\begin{tabular}{|l|l|}
\hline
\textbf{Teacher Model} & \textbf{Student Models} \\ \hline
\multirow{3}{*}{Qwen2.5-7B} 
    & Qwen2.5-3B \\
    & Qwen2.5-1.5B \\
    & Qwen2.5-0.5B \\ \hline
\multirow{5}{*}{Pythia-2.8B} 
    & Pythia-1.4B \\
    & Pythia-1B \\
    & Pythia-410M \\
    & Pythia-160M \\
    & Pythia-70M \\ \hline
\end{tabular}
\caption{Teacher and student model configurations evaluated in our experiments.}
\label{table:teacher_student_models}
\end{table}

\subsection{Variance Analysis}
\label{sec:var_anal}
Table~\ref{tab:std-deviation} reports the mean and standard deviation of EM and ROUGE-L across five runs with different random seeds for distilled student models from both model families. We find consistent results with low variance, which indicates that our findings are stable and reproducible across different training runs.

\begin{table}[th!]
\centering
\resizebox{\columnwidth}{!}{%
\begin{tabular}{l l r r}
\toprule
\textbf{Model} & \textbf{Dataset} & \textbf{EM (±std)} & \textbf{ROUGE-L (±std)} \\
\midrule
Qwen2.5-3B & MLQA-EN (Zero) & 40.33 ± 0.31 & 57.28 ± 0.36 \\
Qwen2.5-3B & MLQA-EN (One) & 48.45 ± 0.76 & 73.47 ± 0.30 \\
Qwen2.5-1.5B & MLQA-EN (Zero) & 39.74 ± 0.49 & 56.45 ± 0.20 \\
Qwen2.5-1.5B & MLQA-EN (One) & 46.97 ± 0.37 & 72.04 ± 0.41 \\
Pythia-1.4B & MLQA-EN (Zero) & 39.15 ± 0.54 & 54.76 ± 0.35 \\
Pythia-1.4B & MLQA-EN (One) & 53.65 ± 0.85 & 75.53 ± 0.63 \\
Pythia-70M & MLQA-EN (Zero) & 13.97 ± 0.53 & 23.90 ± 0.53 \\
Pythia-70M & MLQA-EN (One) & 23.21 ± 0.52 & 34.64 ± 0.39 \\
\bottomrule
\end{tabular}
}
\caption{Variance analysis (mean ± std) for student models across five seeds.}
\label{tab:std-deviation}
\end{table}

Notably, the smallest model in the Pythia family, Pythia-70M, shows low variance across runs, indicating stable behavior. Although its overall performance is lower than larger models, its consistency suggests that performance stability does not necessarily correlate with high accuracy. This highlights that even low-capacity models can produce reliable outputs, further supporting the robustness of our evaluation setup.

\end{document}